\UseRawInputEncoding

\documentclass[10pt,conference,a4paper]{IEEEtran}

\IEEEoverridecommandlockouts
\usepackage{amsmath}
\usepackage{caption}
\usepackage{graphicx, subfig}
\usepackage[nocompress]{cite}
\usepackage{bm}
\usepackage{color}

\ifCLASSINFOpdf
\else
\fi
\begin{document}
%
\title{Open Set Domain Recognition via Attention-Based\\GCN and Semantic Matching Optimization}

\author{\IEEEauthorblockN{Xinxing He, Yuan Yuan, Zhiyu Jiang*}
\IEEEauthorblockA{School of Computer Science and Center for OPTical IMagery Analysis and Learning(OPTIMAL),\\
Northwestern Polytechnical University, Xi'an, Shaanxi, P.R. China\\
hxx583946756@mail.nwpu.edu.cn, y.yuan1.ieee@gmail.com, jiangzhiyu@nwpu.edu.cn}
\thanks{IAPR 2020. Personal use of this material is permitted. Permission from IEEE must be obtained for all other uses, in any current or future media, including reprinting/republishing this material for advertising or promotional purposes, creating new collective works, for resale or redistribution to servers or lists, or reuse of any copyrighted component of this work in other works. *Corresponding author: Zhiyu Jiang (jiangzhiyu@nwpu.edu.cn)}
}

%


\maketitle

\begin{abstract}
Open set domain recognition has got the attention in recent years. The task aims to specifically classify each sample in the practical unlabeled target domain, which consists of all known classes in the manually labeled source domain and target-specific unknown categories. The absence of annotated training data or auxiliary attribute information for unknown categories makes this task especially difficult. Moreover, exiting domain discrepancy in label space and data distribution further distracts the knowledge transferred from known classes to unknown classes. To address these issues, this work presents an end-to-end model based on attention-based GCN and semantic matching optimization, which first employs the attention mechanism to enable the central node to learn more discriminating representations from its neighbors in the knowledge graph. Moreover, a coarse-to-fine semantic matching optimization approach is proposed to progressively bridge the domain gap. Experimental results validate that the proposed model not only has superiority on recognizing the images of known and unknown classes, but also can adapt to various openness of the target domain.

Keywords-open set domain recognition; attention mechanism; semantic matching
\end{abstract}



%
\IEEEpeerreviewmaketitle

\section{Introduction}
With the great progress and wide application of deep learning, the performance of various visual tasks has been improved. However, a large amount of labeled training data are the substantial prerequisite for obtaining a visual task model with excellent performance, which is obviously limited in many real application scenarios. A reasonable solution is to leverage a relevant domain with abundant supervision information to learn something useful for the domain of interest. Samples from different domains may follow different distributions, leading the model trained on the source domain often make false predictions on the target domain. Domain Adaptation (DA) is an effective way to minimize the domain discrepancy and train a better model on the source domain that is successfully applicable to the target domain.

\begin{figure}[!t]
\centering
\includegraphics[width=3.5in]{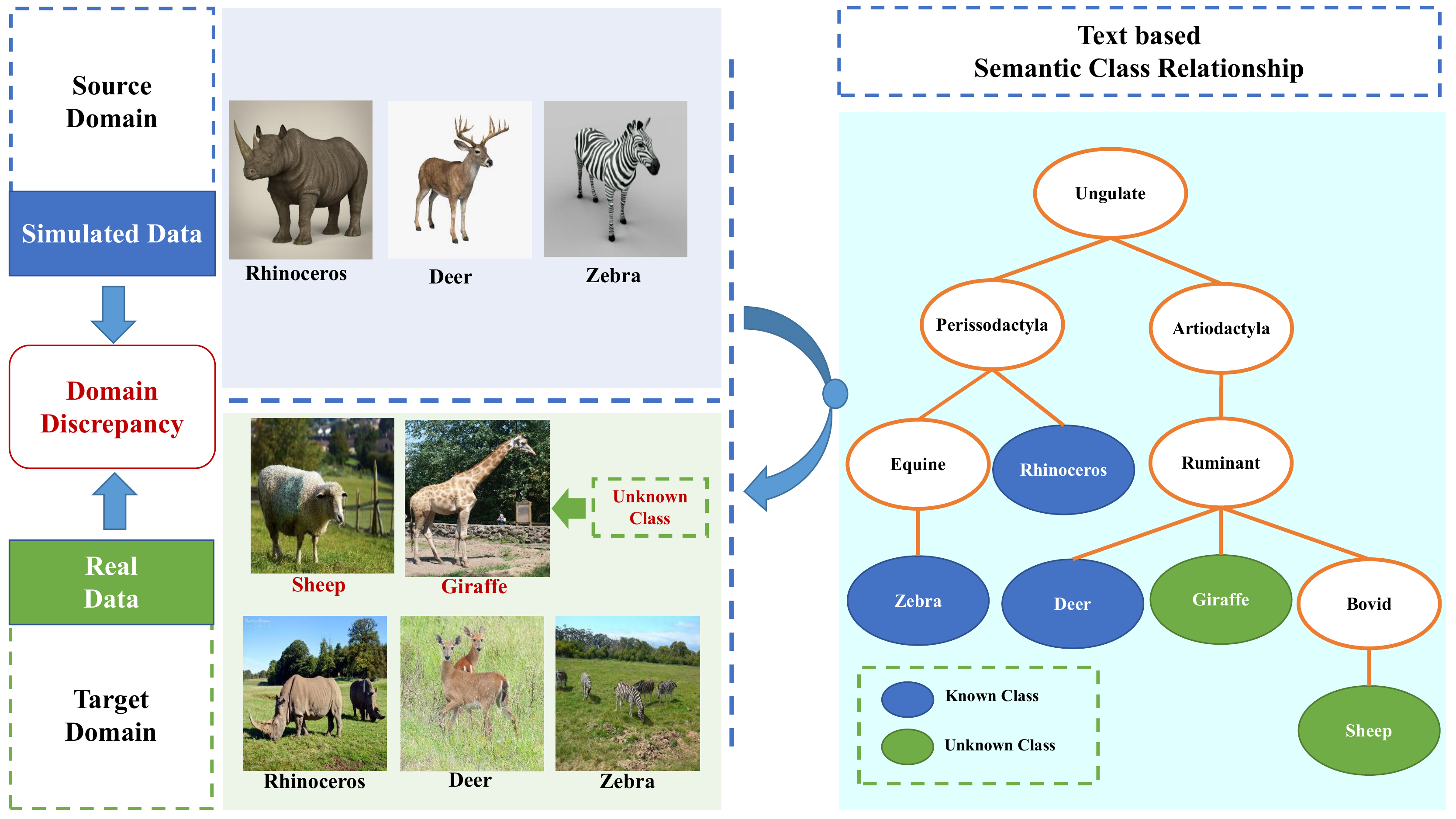}%
\caption{Illustration of the task of Open set domain recognition. The samples of the target domain are taken from natural scene while the examples of the source domain are collected from simulative images which are discrepant from the target data. In addition, the target domain contains unknown categories which are absent in the source domain, and the task is to recognize the specific label of unknown categories with semantic relationships rather than only the ``unknown'' category.}
\label{OSDR}
\end{figure}

Existing methods of domain adaptation can be divided into closed set domain adaptation \cite{DBLP:conf/icpr/RoyBM18,luo2020unsupervised,kang2019contrastive,tzeng2017adversarial} and open set domain adaptation \cite{ryu2020collaborative, panareda2017open, saito2018open, zhuo2019unsupervised}. Compared with closed set domain adaptation which assumes that identical labels are shared across domains, open set domain adaptation is more realistic. For open set domain adaptation, the target domain contains all categories of the source domain and owns target-specific categories. Furthermore, existing open set domain adaptation methods tend to correctly classify samples of known classes in the target domain and divide all samples of unknown classes into ``unknown'' class. In fact, classifying these unknown classes specifically is what we authentic demand in numerous applications. Therefore the Open Set Domain Recognition (OSDR) is studied in this paper. As shown in Fig. \ref{OSDR}, the goal of OSDR task is to identify all samples in the target domain, including known and unknown classes under domain discrepancy. As shown in TABLE \ref{table1}, OSDR is also different from Zero-Shot Learning (ZSL) \cite{DBLP:conf/icpr/RoyBM18,kampffmeyer2019rethinking,song2018transductive,xian2017zero} for the fact that the samples in ZSL are distributed in the same domain and domain gap does not exist. Therefore, the open set domain recognition task is realistic and challenging since it tries to recognize the specific label for unknown classes without training samples.

Open set domain recognition faces two major technical issues. First, it is difficult to specifically classify the unknown categories in the target domain since these samples do not have annotated training data or any auxiliary attribute information. Only the knowledge of the relationship among known and unknown classes can be used to propagate classification regulations of known classes to unknown classes. Therefore, the key idea to deal with the unknown categories is transferring knowledge obtained from familiar classes to describe novel classes. zGCN \cite{wang2018zero} utilizes the word embedding and the categorical relationships encoded by WordNet \cite{miller1995wordnet} to estimate the classifiers of unknown classes. Based on zGCN, Dense Graph Propagation (DGP) \cite{kampffmeyer2019rethinking} is proposed which designs a hierarchical graph structure to enjoy the benefits of the graph structure and prevent knowledge dilution from distant nodes. In \cite{song2018transductive}, Song \emph{et.al} proposed Balance Constraint GCN (bGCN) with the balance constraint loss to prevent known classes from being classified into unknown classes. Although these methods progressively improve the identification precision of the unknown classes by changing network structure or adding additional constraints, most of them ignore that different neighbors contribute differently to the center node in the process of knowledge propagation. Additionally, due to the domain gap between the target domain and the source domain in OSDR, the transferred classification rules of unknown categories may lack the ability to generalize. In order to alleviate the semantic shift, attention-based Graph Convolutional Networks (attention-based GCN) is adopted to transfer information among different categories inspired by \cite{velivckovic2017graph}. Through the attention mechanism, the neighbor nodes with greater influence on the center node are given greater weight, and the unknown classes can be discriminatively described.

Second, the samples of source domain and target domain are quite different in label space and data distribution. In this circumstance, aligning the whole distribution directly, such as Maximum Mean Discrepancies (MMD) \cite{long2015learning} and Deep CORelation ALignment (DCORAL) \cite{sun2016deep}, will lead to seriously negative transfer. The number of unknown categories in the target domain has a great influence on the effect of domain adaptation. None of these models mentioned above, including zGCN, DGP, and dGCN, take the domain discrepancy into account. However, domain discrepancy in OSDR problem makes attention-based GCN transfer semantic embedding to unknown categories in biased manner, which can also degrade the performance of the classifier. In order to minimize the negative impact of domain gap, Semantic Matching Optimization (SMO) method is proposed to make accurate and efficient matching between the target and source domain. If the matched results are accurate enough, the influence brought by unknown classes can be reduced to some extent since the domain discrepancy is measured by the feature distance on these matched pairs.

For tackling the issues described above, the contributions of this work are listed as follows:

1) In order to address the problem that recognition accuracies of unknown classes is low in open set domain recognition, attention-based GCN is applied to transfer information in the knowledge graph. Through the attention mechanism, the neighbors with greater influence on the center node are given greater weight, so that the unknown classes can learn more discriminating feature representations and further obtain more accurate visual classifier.

2) In order to solve the problem that domain discrepancy makes attention-based GCN transfer biased knowledge to unknown categories in OSDR, semantic matching optimization method is designed to perform domain invariant feature learning by minishing the domain gap measured on the shared classes data in the source domain and the target domain.
\begin{table}[!t]\scriptsize
\vspace{1em}
\caption{The major differences among closed set DA, open set DA, ZSL, and OSDR tasks.}
\label{table1}
\centering
\begin{tabular}{cccc}
\hline
& Domain & Unknown Classes & Specific Labels for\\
&Discrepancy & in Target Domain & Unknown Classes\\
\hline
Closed DA & Yes & No & No\\
Open DA & Yes & Yes & No\\
ZSL & No & Yes & Yes\\
OSDR & Yes & Yes & Yes\\
\hline
\end{tabular}
\end{table}

\begin{figure*}[!t]
\centering
\includegraphics[width=6.0in]{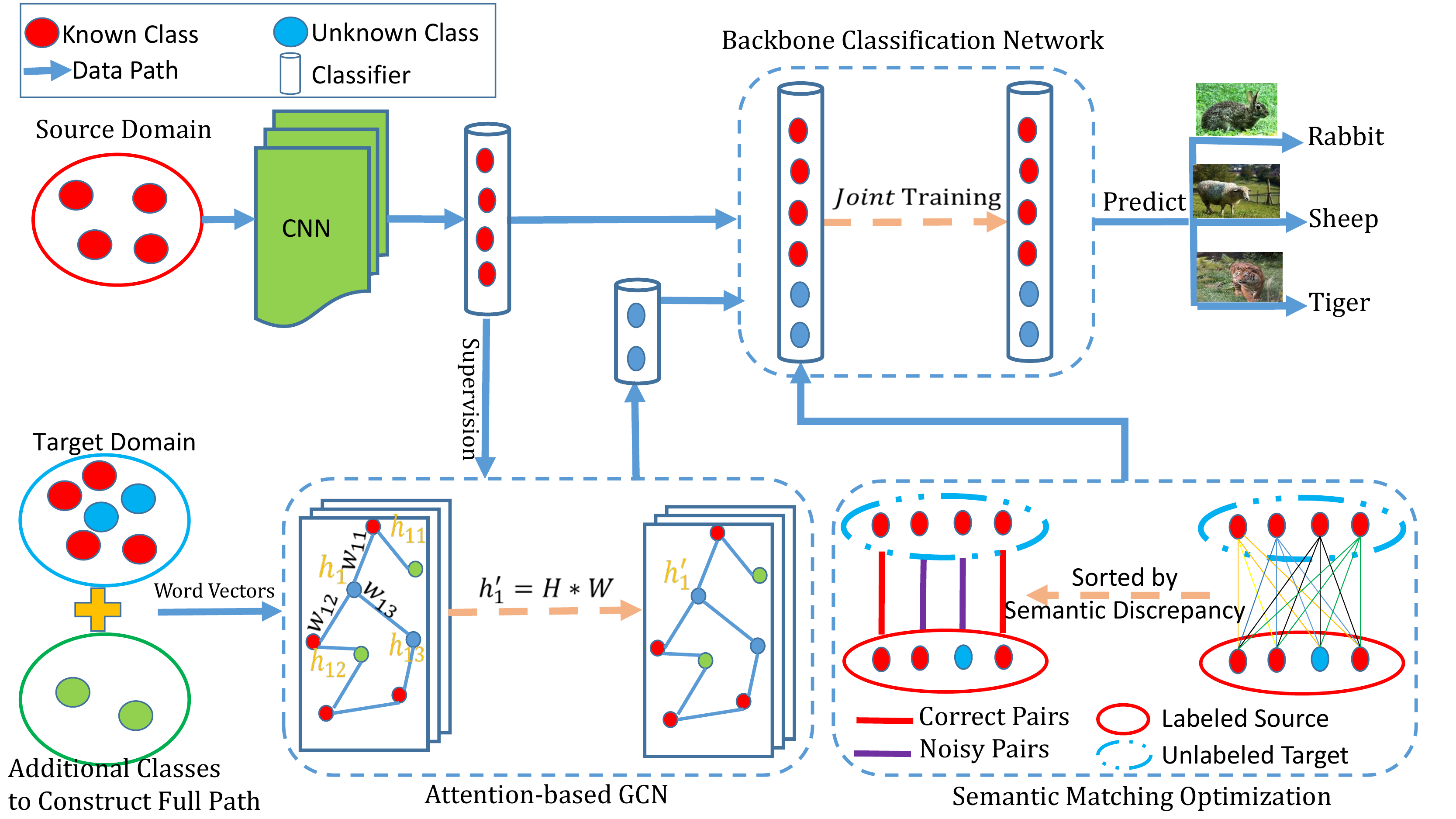}%
\caption{The framework of the proposed open set domain recognition method. Attention-based GCN is utilised to obtain classifier weights of unknown classes. And the backbone classification network is updated based on minimizing the matching discrepancy and transfer loss. Furthermore, the classification network and attention-based GCN are jointly trained in an end-to-end manner.}
\label{network}
\end{figure*}
\section{Related Works}
In this section, previous works related to the proposed method are reviewed.

\textbf{Open Set Domain Adaptation.} Open set domain adaptation is the topic of addressing the domain discrepancy and asymmetric classes between the source domain and the target domain. Recently, several works have been contributed to this issue. Bendale et al. \cite{bendale2016towards} proposed a OpenMax layer to estimate the unknown classes. Ge et al. \cite{ge2017generative} further improved the OpenMax layer by utilizing the GAN to clarify the probability of unknown categories. Assign-and-Transform-Iteratively method \cite{panareda2017open} exploits the feature distance between each target sample and the center of each source category to decide whether a target sample belongs to one of source categories or the ``unknown'' category. Open Set Back-Propagation in \cite{saito2018open} trains a feature generator to guide the probability that a target sample is classified as “unknown” class deviate from the predefined threshold. The feature extractor and classifier are trained in an adversarial training framework. These methods can tackle the domain gap effectively, while all of them only divide the samples of the unknown classes into the “unknown” class without specific classification.

\textbf{Zero-Shot Learning.} As the most relevant problem with OSDR, zero-shot learning (ZSL) task aims to classify novel object instances from unseen categories by learning an embedding space between image and  discriminative semantic representations. However, unlike OSDR, there is no assumed domain gap between the training dataset and the testing dataset. Propagation semantic transfer (PST) \cite{rohrbach2013transfer} utilizes multiple constructs of unknown classes by combining external knowledge for label propagation. Guo et al. \cite{guo2016transductive} proposed the joint learning method for obtaining shared model space of models, in which knowledge can be propagated effectively among categories by using attributes. In \cite{song2018transductive}, Unbias ZSL is proposed to enforce a balanced classifier response for unlabeled target data between known and unknown classes to learn the unbiased embedding space of ZSL.

\textbf{Semantic Embedding and Knowledge Graph.} Early work on zero-shot learning can be divided into two main research directions. The first method is to denote each class with learned semantic representations which can be corresponded to visual classifiers. Socher et al. \cite{socher2013zero} proposed to learn the linear mapping between sample features and word embedding by training two distinct neural networks for images and vector representations in an unsupervised pattern. The second approach is to use knowledge graph. For example, Salakhutdinov et al. \cite{salakhutdinov2011learning} utilized WordNet to encode word vectors among distinct classes such that objects without training samples can learn statistical strength from familiar objects. In \cite{wang2018zero}, semantic embedding and knowledge graph are combined to distill information. Specifically,the model estimates the visual classifier of invisible classes by giving the vector representations of novel categories and knowledge graph which encodes explicit relationships.

\textbf{Attention-Based GCN.} Attention mechanism has proven to be useful in many tasks (Liu et al. \cite{DBLP:conf/icpr/LiuWL0ZHL18}; Ahmad et al. \cite{DBLP:journals/access/AhmadMLT20}). One of the benefits of attention mechanism is that it emphasises on the most relevant parts of the input. Nathani et al. \cite{nathani2019learning} proposed to assign distinct attention values to the neighbor node with different distances from the given node by iterating the attention mechanism network, so that the embedded vector of the node contains multi-hop neighbor node information. In \cite{velivckovic2017graph}, attention-based GCN is applied to perform node classification of  graph-structured data. The method is efficient since it is parallelized across node neighbors. In addition, the architecture can be directly applicable to inductive learning, including tasks where the model has to generalize to unseen graph.

\section{Method}
In this section, attention-based GCN is introduced to address the problem that recognition accuracy of unknown classes is low for the OSDR task, and then we describe the Semantic Matching Optimization (SMO) method which aims to reduce the negative impact of domain discrepancy on knowledge transfer.

Suppose that there are $n_s$ training examples belonging to $l_s$ categories in the source domain. Let $\bm{z}_s$ = ${\{z_i^s}\}_{i=1}^{n_s}$ and $\bm{y}_s$ = ${\{y_i}\}_{i=1}^{n_s}, y_i \in \{1, 2, . . . , l_s\}$ denote  samples and labels respectively. Additionally, the target domain contains $n_t$ unlabeled examples represented as $\bm{z}_t$ = ${\{z_j^t}\}_{j=1}^{n_t}$, and their labels represented as $\bm{y}_t$ = ${\{y_j}\}_{j=1}^{n_t}, y_j \in \{1, 2, . . . , l_t\}$. For the open set domain recognition task, the value of $l_t$ is not available and $l_s < l_t$. That is, there are $l_u$ = $l_t - l_s$ unknown categories in target domain. We denote the feature extractor as ${\varphi}(\cdot)$ and classifier response as ${\psi}(\cdot)$ in this work.

\subsection{Attention-based GCN for Knowledge Propagation}
For the unstructured distribution of knowledge graph, GCN is well designed to leverage the neighborhood information. However, when calculating the hidden feature representations of each node, the GCN assigns the average weight to the neighbor node. Such propagation rules are not capable of capturing which neighbors are more relevant to recognize target nodes, which is necessary in many real applications where edges between different category nodes usually imply different strengths of relations. For example, the giraffe and zebra are two neighbor nodes of the okapi. Most features of the okapi, such as stripes, limbs, ears, are highly similar to the zebra, while only a few features are similar to the giraffe. Therefore, when obtaining the hidden representations of okapi node, the adjacent edge of zebra and okapi should be given more weight. Inspired by this fact, attention-based GCN is proposed to compute the feature representations of each node in the graph by attending over its neighbors following attention strategy.

We first build a graph with $n$ nodes where each node represents a distinct class. Additional category nodes are required to construct complete path from these known classes to unknown classes so that the classification rules can be transferred. Every category node is initialized with a $c$-dimensional semantic vector $\bm{X}$ = \{$ \bm{h}_1$, $\bm{h}_2$, $\dots$, $\bm{h}_n$\}, ${\bm{h}_i} \in {R^c}$. We use $\textbf{Z}$ =\{$\bm{z}_1, \bm{z}_2, \dots, \bm{z}_n$\}, ${\bm{z}_i} \in {R^f}$ denote the output of attention-based GCN. And then, we perform self-attention mechanism on these nodes. The attention coefficient is computed as
\begin{equation}
{\alpha}_{ij} = f_a(\bm{W} \times \bm{h}_i,\bm{W} \times \bm{h}_j),
\end {equation}
where ${\alpha}_{ij}$ is attention coefficient which indicates the importance of node $j$ to node $i$. The attention machinism $f_a(\cdot)$ can be a function of measuring similarity based on cosine distance or Euclidean distance. $\bm{W} \in {R^{f \times c}}$ is weight matrix to convert the original input features into high level representations. After obtaining attention coefficient, the final output features of every node is expressed as

\begin{equation}
\bm{z}_i = \sum_{j \in {n_i}}{{{\alpha}_{ij}} \times \bm{W} \times \bm{h}_j},
\end {equation}
where $n_i$ is a clique that represents first-order neighbors of node $i$ in the graph. In our experiment, the first $m$ categories ($m<n$) have enough data to estimate their classifier weights as ground-truth. And then we can predict classifier weight by equation (2). During training, the mean-square error of the ground-truth and the predicted classifiers is utilized as loss function. By minimizing the loss from the $m$ classes with abundant supervision to estimate the network parameter for the attention-based GCN. The loss is
\begin{equation}
L_{init} = {\frac{1}{m}}{\sum_{i=1}^m{{(\bm{z}_i - \bm{w}_i)}^2}}.
\end {equation}

Finally, we utilize the estimated classifiers to initialize the classification network for the source and target domain.

\subsection{Semantic Matching Optimization}
For the open set domain recognition task, domain discrepancy between the source and target domain exists. Such domain gap makes attention-based GCN propagate biased knowledge to unknown classes, leading performance degradation on the target domain. Hence, minimizing the domain discrepancy is necessary. However, measuring the domain discrepancy is quite difficult in OSDR problem since there exist numerous samples of unknown categories. Zhuo et al. \cite{zhuo2019unsupervised} proposed Hungarian algorithm to generate matching pairs between the two distinct domains and then minimize the domain differences. Experiments show that when using the Hungarian algorithm for matching, the accuracy of the matching is low and the time consumption is high. Since there are more than ten thousand images in the target domain, it is necessary to process a square matrix with dimensions greater than ten thousand when using the Hungarian algorithm for matching, which requires a large amount of computation and high time complexity. If data are grouped, incomplete data will lead to the result that may not be the optimal solution. In this case, the difficulty of filtering noise matching increases, which further augments the negative impact of unknown classes on minimizing domain discrepancy. So we utilize a more appropriate matching method in our model.

For each sample in the target domain, we calculate the feature distance between it and each sample in the source domain. Then the source sample with the minimum distance is selected as its optimal pair. By using this simple and effective method, optimal matched pairs between the two domains can be obtained. Directly minimizing the discrepancy measured via noisy matched instances leads to negative transfer inevitably. Hence, we use semantic consistency to filter such noisy matching. Specifically, $\bm{z}_i^s$ and $\bm{z}_i^t$ is a matched pair. We denote their features as $\bm{f}_i^s = \varphi(\bm{z}_i^s)$ and $\bm{f}_i^t = \varphi(\bm{z}_i^t)$, and calculate their classifier responses $\bm{p}_i^s = \psi(\bm{f}_i^s)$ and $\bm{p}_i^t = \psi(\bm{f}_i^t)$ respectively, the domain discrepancy can be measured as
\begin{equation}
L_{d} = {\sum_{i}{d(\bm{f}_i^s,\bm{f}_i^t)*I}},
\end{equation}

\begin{equation}
I=
\begin{cases}
1, &  d(\bm{p}_i^s,\bm{p}_i^t)<{\tau},\\
0, &  else,
\end{cases}
\end{equation}
where $d(\bm{f}_i^s,\bm{f}_i^t)$ is L2 distance in our work while other distance metrics can also be utilized. $\tau$ is a pre-defined threshold. In fact, if $\bm{z}_i^s$ and $\bm{z}_i^t$ is a right pair, the $d(\bm{p}_i^s,\bm{p}_i^t)$ will be small since samples belonging to the same class are assumed to own highly similar classification response.

\subsection{Network Architecture}
As shown in Fig. \ref{network}, our network architecture which abbreviated as AGCN-SMO consists of an attention-based GCN that maintains the relationship among all classes and a backbone classification network. We first utilize attention-based GCN to generate classifier weights for unknown categories in the target domain. And then the classification layer of backbone network is initialized by these obtained parameters. Based on the initialized classification network, we further joint all proposed approach to fine tune. It is worth noting that the difference between the source and target domain affects the migration of knowledge from unknown to known classes, so we also consider the transfer loss ($L_{tran}$) of attention-based GCN during joint training. The calculation method of $L_{tran}$ is the same as equation ($3$), but the initial input of attention-based GCN is not the word vectors of each category, but the network parameters obtained in section $A$. So the total loss is
\begin{equation}
Loss = L_{cls} + L_{tran} + L_{lb} + L_d,
\end{equation}
where $L_{cls}$ is classification loss on the labeled source data. $L_{lb}$ is limited balance constraint proposed in \cite{song2018transductive} which aims to prevent samples of unknown classes being recognized as known classes in the target domain.
\section{Experiments}
\subsection{Dataset Description and Evaluation Metrics}
We verified our method on I2AwA \cite{zhuo2019unsupervised} and I2CIFAR dataset. The target domain of I2AwA is AwA2 \cite{xian2017zero} which contains 50 animal classes. There are altogether 37,322 images in AwA2, that is, an average of 746 images per category. In order to consider the impact of openness (the proportion of unknown categories in total classes) on OSDR problem, we divide the target domain into two different settings to perform experiment:
\begin{itemize}
  \item $openness=0.2$, where 40 known categories and 10 unknown categories;
  \item $openness=0.4$, where 30 known categories and 20 unknown categories.
\end{itemize}

Zhuo et al. \cite{zhuo2019unsupervised} collected a dataset with 2970 simulated images belonging to 40 classes via Google image searching engine. According to the experimental settings mentioned above, we respectively select 40 and 30 categories as the data of the source domain. Obviously, there is a large domain gap between the source and target domain.

As for I2CIFAR, the source domain contains 1130 simulated images belonging to 15 classes and the target domain consists of 45 classes in the CIFAR \cite{DBLP:conf/nips/OreshkinLL18} dataset. That means there are 30 unknown classes and 15 known classes in the target domain.

We perform classification on the whole target domain. The evaluation metrics are the Top1 Classification Accuracies of known categories, unknown categories and all categories on target domain. The accuracies of known and unknown categories are set to show the influence of the proposed method on the identification of known and unknown classes more intuitively. The accuracies of all categories are set for better understanding the effectiveness of the method as a whole.

\subsection{Implementation Details}
We construct two knowledge graphs based on the WordNet \cite{miller1995wordnet} for distinct settings. The nodes in these graphs include the whole classes in the target domain and the descendants and ancestors of them. To extract the word embedding as the input of attention-based GCN, we utilize the GloVe text model \cite{pennington2014glove} trained on the Wikipedia dataset, which leads to 300-d vectors. ResNet-50 \cite{he2016deep} pretrained on ImageNet is used as basic model where the parameters of the last classifier layer is taken as the aim that attention-based GCN tends to estimate. The attention-based GCN model is composed of two layers. The first layer consists of two attention heads to compute hidden representations. The second layer with a single attention head is applied for obtaining the final output whose dimension depends on the parameters of the last classification layer in the ResNet-50 architecture. The supervision for training attention-based GCN is classifiers optimized on the annotated source domain. And then we concatenate these initialized classifiers to the pre-trained ResNet-50 whose original classification layer is removed to form the classification network for the source domain and the target domain. The Semantic Matching Optimization (SMO) method is designed to get optimal matched pairs for estimating and further reducing the domain discrepancy. Finally, transfer loss, balance constraint loss, classification loss and matching loss are combined to train the backbone classification network.

\begin{table}[!t]
\caption{Classification accuracy on I2AwA dataset, where $openness=0.2$.}
\label{openness=0.2}
\centering
\begin{tabular}{cccc}
\hline
& Known & Unknown  & all\\
\hline
zGCN & 77.2 & 21.0 & 65.0\\
dGCN & 78.2 & 11.6 & 64.0\\
adGCN & 77.3 & 15.0 & 64.1\\
bGCN & 84.6 & 28.0 & 72.6\\
pmd-bGCN & 84.7 & 27.1 & 72.5\\
UODTN & 84.7 & 31.7 & 73.5\\
AGCN-SMO(ours) & \textbf{85.1} & \textbf{34.3} & \textbf{74.3}\\
\hline
\end{tabular}
\end{table}

\begin{table}[!t]
\caption{Classification accuracy on I2AwA dataset, where $openness=0.4$.}
\label{openness=0.4}
\centering
\begin{tabular}{cccc}
\hline
& Known & Unknown  & all\\
\hline
zGCN & 69.7 & 13.2 & 45.3\\
dGCN & 74.3 & 4.3 & 43.9\\
adGCN & 71.9 & 7.1 & 43.8\\
bGCN & 77.8 & 19.7 & 52.6\\
pmd-bGCN & 79.6 & 19.9 & 53.7\\
UODTN & 77.9 & 24.1 & 54.5\\
AGCN-SMO(ours) & \textbf{79.7} & \textbf{27.4} & \textbf{56.7}\\
\hline
\end{tabular}
\end{table}

\begin{table}[!t]
\caption{Classification accuracy on I2CIFAR dataset.}
\label{cifar=0.4}
\centering
\begin{tabular}{cccc}
\hline
& Known & Unknown  & all\\
\hline
zGCN & 50.1 & 5.6 & 21.7\\
dGCN & 59.9 & 3.1 & 22.1\\
adGCN & 59.3 & 4.2 & 22.6\\
bGCN & 65.3 & 9.3 & 27.9\\
pmd-bGCN & 63.9 & 11.8 & 29.2\\
UODTN & 64.2 & 13.6 & 30.4\\
AGCN-SMO(ours) & \textbf{65.1} & \textbf{15.7} & \textbf{32.1}\\
\hline
\end{tabular}
\end{table}

\subsection{Results and Discussion}
We implement the comparative experiments with following baselines: zGCN \cite{wang2018zero}, two improvements including dGCN and adGCN proposed in \cite{kampffmeyer2019rethinking}, bGCN \cite{song2018transductive} and pmd-bGCN \cite{chen2017population}. zGCN utilizes the word vectors extracted  by the GloVe text model \cite{pennington2014glove} and the categorical relationship encoded in the WordNet to estimate the classifiers of unknown classes. Based on zGCN, the authors in \cite{kampffmeyer2019rethinking} design a hierarchical graph structure (dGCN) and further assign distinct weights to different hops away from given node (adGCN). Specifically, adGCN assigns the same weight to nodes which are the same distance from the center node.   bGCN propose original balance constraint in state-of-the-art transductive zero-shot learning methods. Furthermore, following bGCN, pmd-bGCN is proposed to reduce population matching discrepancy which shows superiority over other domain discrepancy measurements. Except these methods, Attention-Based GCN and Semantic Matching Optimization (AGCN-SMO) model is also compared with Unsupervised Open Domain Transfer Network (UODTN) \cite{zhuo2019unsupervised}, which is the fastest network on OSDR problem. As shown in TABLE \ref{openness=0.2} and TABLE \ref{openness=0.4}, we can obtain the following observations: 1) AGCN-SMO outperforms all the baselines to some extent. zGCN, dGCN and adGCN can not adapt well to target data since the classification confusion between these known and unknown classes is severe. UODTN and bGCN display improvements indicating that a balanced constraint on the classifier responses leads to better generalization of model. UODTN is superior to bGCN means reducing domain discrepancy between the source domain and the target domain is efficient. Compared with UOTTN, AGCN-SMO not only utilizes the attention mechanism to learn more discriminative feature representations for unknown classes, but also design SMO to further reduce the negative impact of domain gap, therefore, AGCN-SMO achieve improvements of 1.8\%, 3.3\% and 2.2\% respectively on known, unknown and all classes when openness is 0.4. 2) When the proportion of unknown classes increased from 0.2 to 0.4, the accuracy of AGCN-SMO on unknown classes decrease by 6.9\%, while the accuracy of other methods decreased by more than 7. This result means that AGCN-SMO is more robust to openness change to some extent. 3)The experimental results of AGCN-SMO on I2CIFAR dataset are not as good as those on I2AwA. On the one hand, since CIFAR dataset is composed of several super-classes, each of which varies greatly, and on the other hand, unknown classes in the target domain leads a large proportion. Nevertheless, the overall classification accuracy is improved relative to other methods.

\begin{figure}[!t]
\centering
\includegraphics[width=4.0in]{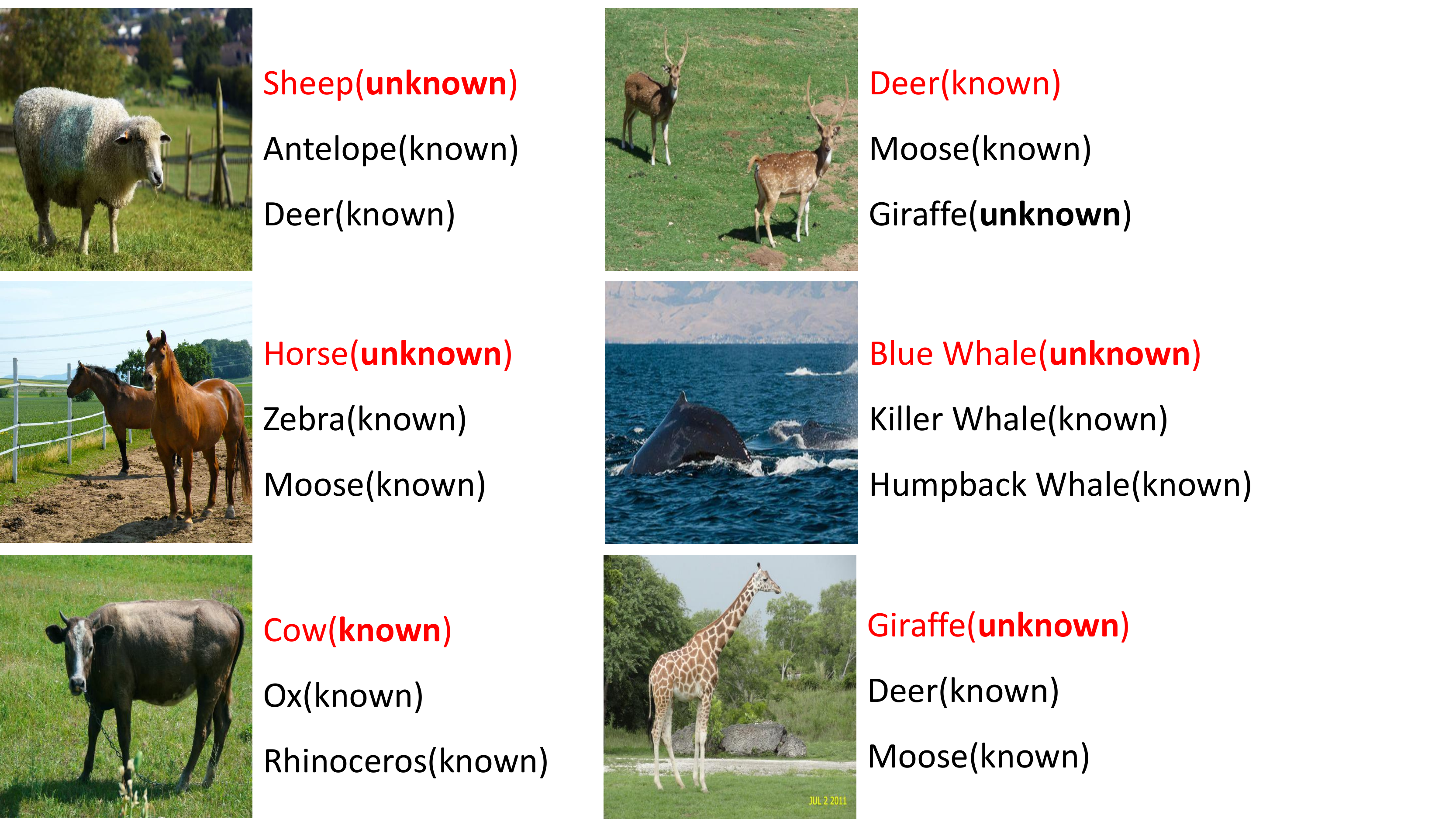}%
\caption{Visualization of top 3 prediction results for various image inputs. The correct predictions are highlighted in red bold characters. The known and unknown classes are marked respectively with ``known'' and bold ``unknow'' in the bracket.}
\label{visualization}
\end{figure}

The visualization of top 3 prediction results for various image inputs is implemented in this paper. As shown in Fig. \ref{visualization}, except the true category to which the sample belongs, the classifier of the more correlated category is also activated with a greater confidence, which indicates the knowledge of the known categories can be transferred well to unknown categories. This property mainly relies on improving the recognition ability of classifiers through attention-based GCN while maintaining semantic consistency with SMO. Another conclusion can be observed that the classification task on this dataset is difficult since it contains several fine-grained classes, such as different kinds of dog, rat, whale and so on. This is also another indication of the effectiveness of the AGCN-SMO as it improves accuracy by 3\% on unknown classes.


\begin{figure}[!t]
\centering
\includegraphics[width=3.5in]{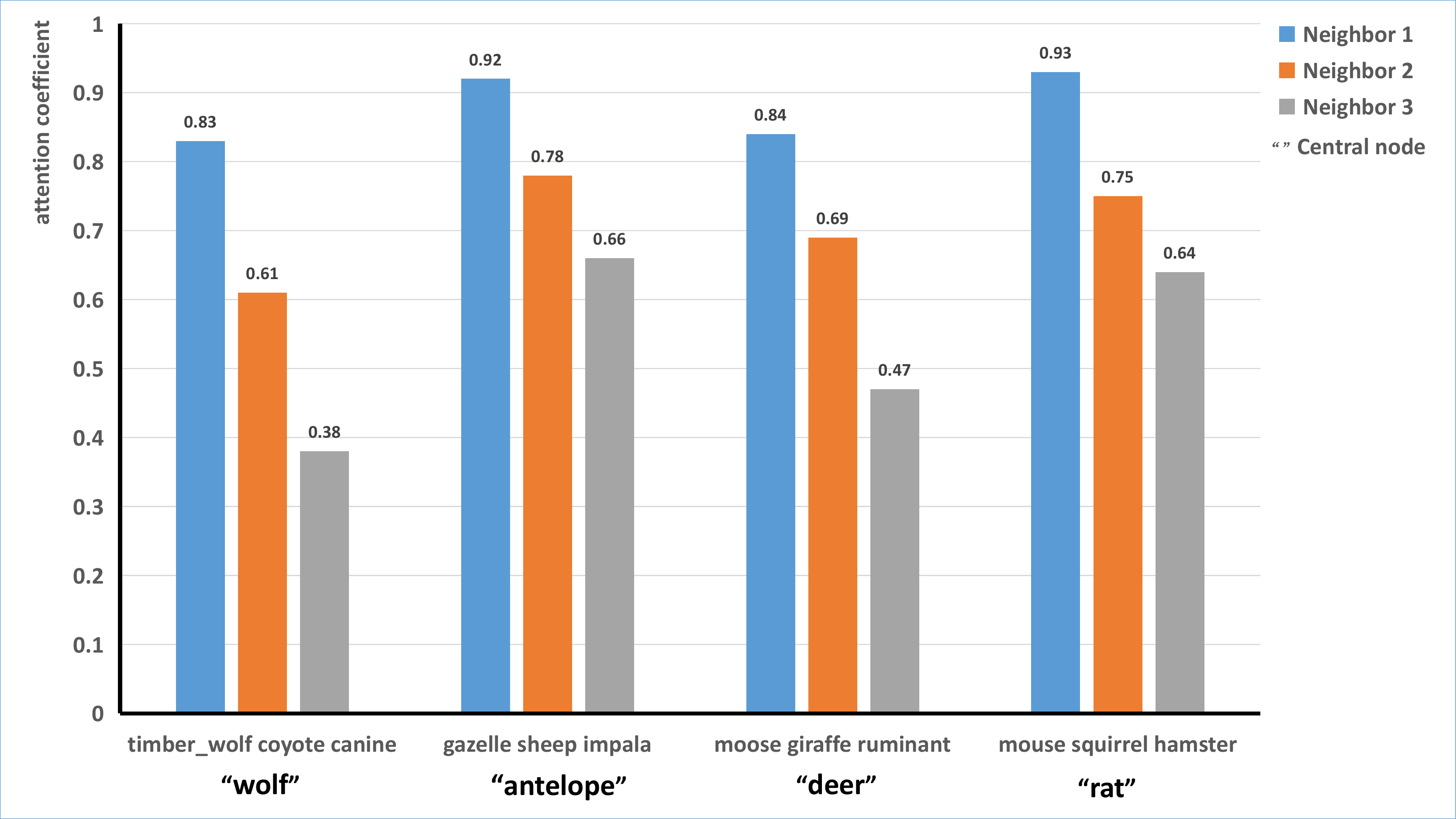}%
\caption{Visualization of top 3 attention coefficient between various central category node and their neighbors. The central category node are highlighted in bold characters.}
\label{weight}
\end{figure}

\subsection{Ablation Study}
As shown in TABLE \ref{ablation study}, we conduct ablation study on I2AwA where openness is 0.2 by evaluating several models to go deeper with the efficacy of Attention-Based GCN and Semantic Matching Optimization(AGCN-SMO). (1) zGCN \cite{wang2018zero} is a basic model without adding any proposed techniques in AGCN-SMO; (2) zGCN+SMO model is performed to validate the  effectiveness of SMO module; (3) The attention-based GCN model adds an attention mechanism on the basis of zGCN; (4) AGCN-SMO is the full model with attention-based GCN and SMO. First, we can see that zGCN+SMO outperforms zGCN by a large margin, which validates that SMO method can effectively avoid negative transfer and learn more domain-invariant features. Further, attention-based GCN improves the performance by 2.4\% on unknown categories and 1.7\% on known categories compared with zGCN. This improvement means that the attention mechanism is efficient for open set domain recognition problem. Fig. \ref{weight} displays top 3 attention coefficient of several central category nodes and their neighbors. The neighbors with greater influence on the center category node are given greater weight via attention mechanism, so that the unknown classes can learn more discriminating feature representations and further obtain more accurate visual classifier. By further integrating attention-based GCN and SMO for joint training, the performance of AGCN-SMO model gains great improvement over other method. It is reasonable as the domain discrepancy affect the migration of classification rules from known classes to unknown classes. By reducing the domain gap during joint training, the transfer loss can be reduced to a certain extent and the classification accuracy can be improved.

\begin{table}[!t]
\caption{Ablation study on I2AwA dataset, where $openness=0.2$.}
\label{ablation study}
\centering
\begin{tabular}{cccc}
\hline
& Known & Unknown  & all\\
\hline
zGCN   & 77.2 & 21.0 & 65.0\\
zGCN+SMO & 81.3 & 27.3 & 69.8\\
Attention-Based GCN      & 78.9 & 23.4 & 67.1\\
AGCN-SMO(ours)  & \textbf{83.1} & \textbf{32.7} & \textbf{72.4}\\
\hline
\end{tabular}
\end{table}

\section{Conclusion}
We have proposed the Attention-Based GCN and Semantic Matching Optimization (AGCN-SMO) network architecture for open set domain recognition task, which conducts joint training between the attention-based GCN and the classification network by minimizing domain difference, reducing classification loss, minimizing transfer loss and limiting balanced constraint loss. Specifically, we employ the attention-based GCN model to learn more distinguishing feature representations for the nodes corresponding to distinct categories from their neighbors which are given different weights in the knowledge graph to solve the problem of low recognition accuracy of unknown classes. Furthermore, the coarse-to-fine SMO module is designed to avoid negative transfer by reducing the domain discrepancy. Experimental results which implemented on I2AwA dataset with various openness show that the proposed model is effective on classifying samples of both known and unknown classes in the target domain. Ablation results have verified the contributions of each component in AGCN-SMO.







%


\bibliographystyle{IEEEtran}
\small{\bibliography{reference}}

\end{document}